\documentclass[journal]{IEEEtran}
\usepackage{ifpdf}
\usepackage{color}
\usepackage{amssymb}
\usepackage[linesnumbered,ruled,vlined]{algorithm2e}
\usepackage{cite}
\usepackage{amsmath}
\usepackage{units}
\usepackage{array}
\usepackage{fixltx2e}
\usepackage{stfloats}
\usepackage{url}
\usepackage{hyperref}
\usepackage{subfigure}
\usepackage{footnote}
\usepackage{multirow}
\usepackage{booktabs}
\usepackage{tabularx}

\ifCLASSINFOpdf
  \usepackage[pdftex]{graphicx}
\else
\fi

\begin{document}
\title{TeLL-Drive: Enhancing Autonomous Driving with Teacher LLM-Guided Deep Reinforcement Learning}

\author{Chengkai~Xu,
        Jiaqi~Liu, 
        Shiyu~Fang,
        Yiming~Cui,
        Dong~Chen,
        Peng~Hang,~\IEEEmembership{Senior Member,~IEEE,}
        and~Jian~Sun 

\thanks{This work was supported in part by the National Natural Science Foundation of China (52302502, 62433014), the Shanghai Scientific Innovation Foundation (No.23DZ1203400), and the Fundamental Research Funds for the Central Universities.}
\thanks{C. Xu, J. Liu, S. Fang, Y. Cui, P. Hang and J. Sun are with the College of Transportation, Tongji University, Shanghai 201804, China. (e-mail: xck1270157991@gmail.com, {liujiaqi13, 2111219, 2301796, hangpeng, sunjian}@tongji.edu.cn)}
\thanks{D. Chen is with the Department of Electrical and Computer Engineering, Michigan State University, Lansing, MI, 48824, USA. (e-mail: chendon9@msu.edu)}
\thanks{Corresponding author: Peng Hang}}


\maketitle

\begin{abstract}
Although Deep Reinforcement Learning (DRL) and Large Language Models (LLMs) each show promise in addressing decision-making challenges in autonomous driving, DRL often suffers from high sample complexity, while LLMs have difficulty ensuring real-time decision making. To address these limitations, we propose TeLL-Drive, a hybrid framework that integrates a Teacher LLM to guide an attention-based Student DRL policy. By incorporating risk metrics, historical scenario retrieval, and domain heuristics into context-rich prompts, the LLM produces high-level driving strategies through chain-of-thought reasoning. A self-attention mechanism then fuses these strategies with the DRL agent’s exploration, accelerating policy convergence and boosting robustness across diverse driving conditions. The experimental results, evaluated across multiple traffic scenarios, show that TeLL-Drive outperforms existing baseline methods, including other LLM-based approaches, in terms of success rates, average returns, and real-time feasibility. Ablation studies underscore the importance of each model component, especially the synergy between the attention mechanism and LLM-driven guidance. Finally, we build a virtual-real fusion experimental platform to verify the real-time performance, robustness, and reliability of the algorithm running on real vehicles through vehicle-in-loop experiments. 
Full validation results are available on
\href{https://perfectxu88.github.io/TeLL-Drive.github.io/}{Our Website}.
\end{abstract}


\IEEEpeerreviewmaketitle
\section{Introduction}
\IEEEPARstart Autonomous driving technology has made significant advancements over the past decade, emerging as a transformative force poised to revolutionize the transportation sector\cite{chen2022milestones, tampuu2020survey}. By promising enhanced safety, reduced traffic congestion, and increased mobility accessibility, autonomous vehicles (AVs) are set to redefine the landscape of modern transportation. Central to the operational efficacy of AVs is their ability to perform real-time, complex decision-making that rivals or surpasses human driving capabilities. Achieving such sophisticated decision-making necessitates the integration of advanced artificial intelligence methodologies capable of perceiving, interpreting, and responding to dynamic and often unpredictable driving environments\cite{aradi2020survey}.


Deep Reinforcement Learning (DRL) has emerged as a key framework for autonomous decision-making\cite{haydari2020deep,liu2023mtd}, with its ability to develop policies for complex tasks such as navigation through intersections \cite{li2022decision, qiao2018automatically} and ramp merging \cite{bouton2019cooperation, wang2017formulation}. DRL’s strength lies in its capacity to learn from experience and optimize driving policies based on trial and error. However, despite its potential, traditional DRL methods face several challenges, including high data demands, slow convergence rates, and limited generalization across diverse and dynamic driving environments \cite{kiran2021deep}. These limitations hinder the scalability and efficiency of DRL in real-world autonomous driving tasks, where adaptability, safety, and real-time performance are paramount.

In parallel, Large Language Models (LLMs), exemplified by architectures such as GPT-4o\cite{hurst2024gpt}, have demonstrated exceptional proficiency in natural language understanding and contextual reasoning. Leveraging vast repositories of knowledge and advanced contextual reasoning capabilities, LLMs can provide valuable insights for decision-making processes in autonomous driving systems\cite{yang2023llm4drive, cui2024survey, xu2024drivegpt4, cui2023drivellm}. However, the deployment of LLMs as standalone decision-making agents faces significant barriers\cite{cui2024survey}. Specifically, LLMs struggle to ensure real-time responsiveness and exhibit a degree of randomness in their decision outputs, which are critical limitations in time-sensitive and safety-critical applications inherent to autonomous driving systems.


To address the limitations, we propose \textbf{TeLL-Drive}, a novel framework that synergistically combines the strengths of both DRL and LLMs to enhance decision-making in autonomous vehicles. By leveraging the contextual understanding and reasoning capabilities of LLMs, TeLL-Drive enhances the sampling efficiency and quality of DRL, while mitigating the data inefficiency and slow convergence typically associated with DRL. Specifically, we introduce a risk-aware LLM agent, equipped with memory, reflective, and reasoning capabilities, that provides context-sensitive guidance to the DRL agent. This enables safer, more efficient decision-making in complex and dynamic traffic scenarios. Meanwhile, the DRL agent, built on the Actor-Critic architecture, ensures robust exploration and real-time responsiveness by employing hybrid strategies, addressing the inherent randomness in LLM-driven decisions.

\begin{figure}
    \centering
    \includegraphics[width=0.9\linewidth]{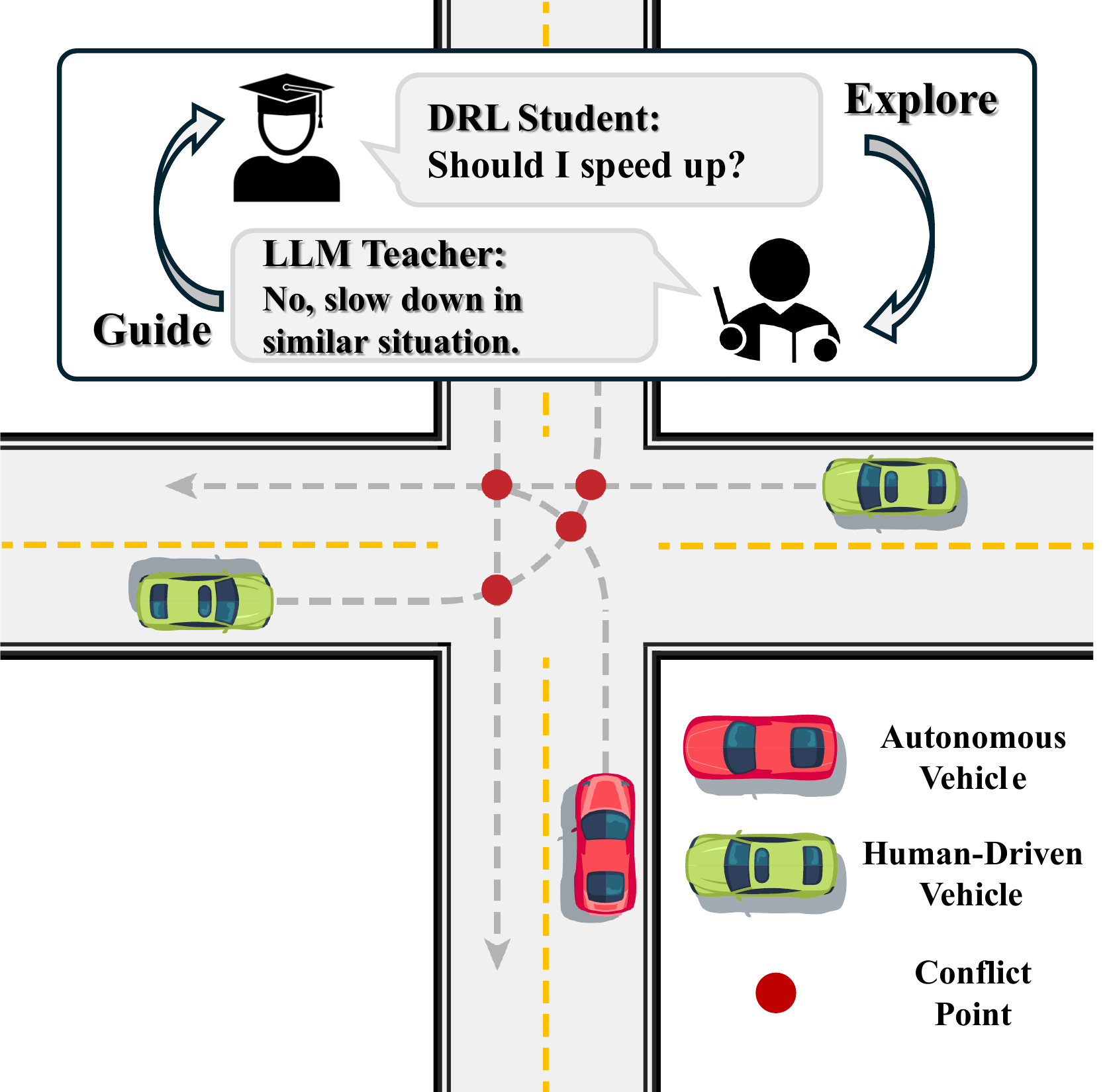}
    \caption{The LLM teacher guides the DRL agent in decision-making within complex traffic scenarios, offering corrective feedback during exploration to enhance learning efficiency and decision-making accuracy.}
    \label{fig:teacher_and_student}
\end{figure}

As shown in Fig.~\ref{fig:teacher_and_student}, the LLM function as a teacher, providing  expert-level guidance and contextual insights that inform and streamline the learning process of DRL agents. Subsequently, DRL serves as the ``student", acting as the final decision maker to ensure real-time responsiveness and mitigate the randomness associated with LLM-driven decisions. The main contributions of this article are listed as follows:

\begin{itemize}
    \item[1)]  We introduce TeLL-Drive, a decision-making framework for autonomous driving that combines a teacher LLM with a student DRL agent, which integrates the LLM’s high-level reasoning and knowledge with DRL’s adaptability and computational efficiency.
    \item[2)] A risk-aware LLM agent is developed, which is endowed with memory, reflection, and reasoning capabilities, to provide context-sensitive guidance in dynamic traffic environments and enhance driving safety and efficiency.
    \item[3)] Through real-vehicle experiments in multiple scenarios, TeLL-Drive outperforms standard DRL algorithms in exploration efficiency and achieves favorable overall results compared to alternative methods.
\end{itemize}

\section{Related Works}
\subsection{DRL for Autonomous Driving Decisions}
DRL has emerged as a promising approach for autonomous driving, spanning tasks from basic lane-keeping to complex multi-agent interactions \cite{liu2024cooperative, xu2018reinforcement}. DRL algorithms, both policy-gradient-based and value-based, have demonstrated substantial performance improvements in simulated driving environments \cite{schulman2017proximal, mnih2013playing}. These methods, by learning from interactions with the environment, can develop highly effective policies for tasks such as intersection navigation \cite{li2022decision}, obstacle avoidance, and adaptive cruise control. However, two main challenges persist in applying DRL to autonomous driving.

First, DRL’s reliance on extensive environment interactions often leads to high data requirements, which can be both costly and time-consuming, particularly when training agents for complex driving tasks. This not only limits scalability but also makes real-world deployment more challenging \cite{kiran2021deep}. Second, DRL models generally lack transparency and interpretability, which impedes their ability to make reliable decisions in rare or out-of-distribution scenarios \cite{zhu2021survey}. This lack of transparency makes DRL less reliable for safety-critical applications, such as handling unexpected or unfamiliar traffic situations.

To address these issues, the integration of expert knowledge through Reinforcement Learning from Human Feedback (RLHF) has been proposed \cite{sun2023aligning}. RLHF allows for faster convergence and improved robustness by leveraging human expertise to guide the learning process, reducing the number of required interactions with the environment. However, RLHF comes with its own set of challenges. First, it is resource-intensive due to the need for extensive human annotations \cite{yu2024rlhf}. Additionally, the human feedback may not cover the full range of possible driving scenarios, limiting the agent's ability to generalize effectively to unseen situations. These limitations point to the need for a more efficient and scalable method that integrates expert guidance while addressing DRL's inherent drawbacks.

\subsection{LLMs in Decision-Making}
LLMs have shown considerable promise in various high-level decision-making tasks, including autonomous driving. LLMs, such as GPT-4, have demonstrated their ability to handle complex reasoning, interpretation, and contextual awareness \cite{fang2024towards}. For example, LanguageMPC \cite{sha2023languagempc} leverages the common-sense reasoning capabilities of LLMs to guide Model Predictive Control (MPC) parameters for autonomous vehicles. Similarly, Fu et al. \cite{fu2024drive} and Wen et al. \cite{wen2023dilu} have explored the application of LLM-based reasoning, interpretation, and memory capabilities to assist autonomous decision-making, particularly in complex and dynamic traffic environments. These models help in interpreting driving scenarios and proposing context-aware strategies based on learned knowledge.

Despite these promising developments, the practical deployment of LLMs in autonomous driving faces several limitations. One of the key challenges is the high computational cost associated with running LLMs in real-time, making it difficult to meet the responsiveness required for safety-critical applications \cite{cui2024survey}. Additionally, LLMs typically generate outputs with a degree of randomness, which can result in unpredictable actions that are unsuitable for tasks demanding consistent and reliable decision-making. This unpredictability is particularly problematic in autonomous driving, where even minor deviations from expected behavior can have serious safety implications. Thus, while LLMs offer significant potential for decision-making in autonomous driving, their practical use as standalone decision-making agents is limited by their real-time performance and output consistency.

\subsection{Hybrid DRL-LLM Approaches}

With the rapid development of LLMs and DRL in various fields, researchers are increasingly exploring the synergistic potential of combining these two paradigms. While numerous studies have focused on using DRL methods to optimize and fine-tune LLMs to enhance their generative capabilities and task adaptability, the utilization of LLMs to assist DRL remains relatively underexplored, particularly in the context of autonomous driving decision-making.

Existing research has begun to investigate how the reasoning and knowledge capabilities of LLMs can improve the exploration efficiency and learning effectiveness of RL agents\cite{liu2024language}. For example, Zhang et al.\cite{zhang2024large} developed a semi-parametric RL framework based on LLMs by configuring long-term memory modules; Similarly, Trirat et al.\cite{trirat2024automl} employed LLMs to achieve full-process automated machine learning, while Ma et al.\cite{ma2023eureka} realized the automatic design of reward functions in RL without requiring specific enhancement tasks. Despite these advancements, the environmental understanding capabilities of LLMs are still not fully leveraged, and effective integration between LLMs and RL remains a challenge. Current approaches lack a comprehensive methodology for combining the strengths of both LLMs and RL, resulting in an under-utilized potential to improve decision-making processes in autonomous driving systems.

\section{Problem formulation}
We formalize the autonomous driving decision-making task as a Partially Observable Markov Decision Process (POMDP), defined by the tuple 
\(\langle \mathcal{S}, \mathcal{A}, \mathcal{O}, \mathcal{T}, \mathcal{R}, \gamma \rangle\), where \(\mathcal{S}\) is the environmental states; \(\mathcal{A}\) is the action space; \(\mathcal{O}\) is the observation space; \(\mathcal{T}\) is the state transition function; \(\mathcal{R}\) is the reward function, and \(\gamma\) is the discount factor. 
The agent’s objective is to learn a policy \(\pi\) that maximizes the expected discounted return:
\begin{equation}
    \max_\pi J(\pi) = \text{arg}\max_\pi \mathbb{E}_{(s_t,a_t)\sim\rho_\pi} [\sum_{t=0}^{\infty} \gamma^t\, r(s_t,a_t)]
\end{equation}
where \(\gamma \in [0,1]\) balances the emphasis on immediate and future rewards.

\subsubsection{Observation space}
At each time step \(t\), the agent receives an observation \(o_t \in \mathcal{O}\) composed of two parts. The first is a matrix \(M_t \in \mathbb{R}^{\mathcal{F}_k \times N}\) capturing information about up to \(N\) nearby vehicles. Each column of \(M_t\) corresponds to one vehicle, described by a feature vector:
\begin{equation}
    \mathcal{F}_k = [x_k,\, y_k,\, v_{x_k},\, v_{y_k},\, \cosh(\theta_k),\, \sinh(\theta_k)]
\end{equation}
where \((x_k,\, y_k)\) and \((v_{x_k},\, v_{y_k})\) denote the position and velocity of the \(k\)-th vehicle, and \(\cosh(\theta_k)\) and \(\sinh(\theta_k)\) encode its orientation. The second part of \(o_t\) is the state of the ego vehicle. By concatenating these components, the agent obtains a compact yet informative representation of the driving environment. 
    
\subsubsection{Action space} This work focuses on leveraging LLMs to provide high-level guidance for DRL, rather than controlling low-level vehicle dynamics. Consequently, the action space \(\mathcal{A}\) consists of five high-level maneuvers:
\begin{small}
    \begin{equation}
    \mathcal{A} = \{slow down, cruise, speed up, turn left, turn right\}
\end{equation}
\end{small}

Once a high-level maneuver is chosen, the corresponding steering and throttle commands are generated by a lower-level PID controller, enabling the vehicle to execute lateral and longitudinal movements. 

\section{Methodology}
\subsection{Framework overview}

\begin{figure*}
    \centering
    \includegraphics[width = 0.9\textwidth]{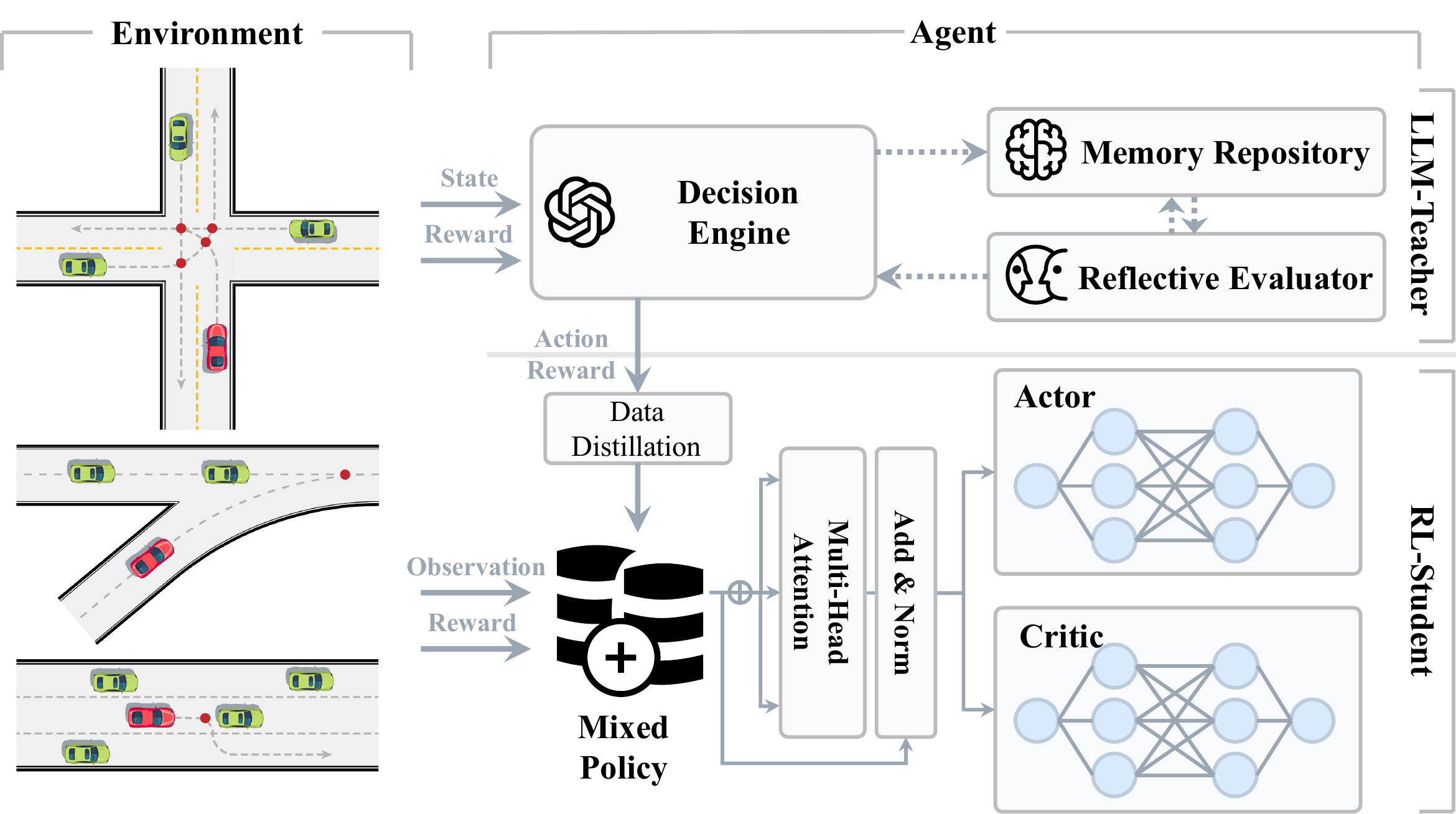}
    \caption{The overall conceptual framework of TeLL-Drive, where a DRL student agent is guided by the LLM teacher for better decision making in autonomous driving.} 
    \label{fig-framework}
\end{figure*}

TeLL-Drive leverages the prior knowledge of LLMs to guide the exploration and learning of DRL agents in  diverse, complex traffic scenarios. By introducing policy integration, TeLL-Drive enhances sample efficiency and optimizes learning outcomes. As illustrated in Fig.~\ref{fig-framework}, the framework comprises two main components: the \textit{LLM Teacher} and the \textit{DRL Student}. 
Based on multi-module collaboration, the Teacher Agent generates robust decision through its three key modules: \textit{Decision Engine}, which provides real-time guidance; \textit{Memory Repository}, which stores past experiences for context; and \textit{Reflective Evaluator}, which refines the guidance based on previous performance.
While the Student Agent refines the Teacher’s actions through a multi-head attention-based policy-integration mechanism, integrating its own exploration experiences to effectively acquire knowledge from the LLM and enhance learning efficiency and quality.

\subsection{LLM Teacher}
\subsubsection{\textbf{Decision Engine}}
The Decision Engine begins by estimating the Time to Conflict Point (TTCP) ~\cite{fang2024towards} for each potential collision, using a rotation-projection method that projects the relative motion vectors of the ego vehicle and other traffic participants onto a shared reference axis.
Let \(\mathbf{d}_\text{ego}(t)\) and \(\mathbf{d}_\text{other}(t)\) be the positions of the ego and another vehicle at time \(t\), \(v_\text{ego}(t)\) and \(v_\text{other}(t)\) be the current speed. The TTCP \(\tau\) is:
\begin{equation}
    \tau = \arg \min_{t \ge 0} \bigl\| \frac{\mathbf{p}_\text{ego}(t)}{v_\text{ego}(t)} - \frac{\mathbf{p}_\text{other}(t)}{v_\text{other}(t)}\bigr\|
\end{equation}

This risk metric informs immediate maneuver priorities. Simultaneously, we retrieve context from a memory repository indexing historical driving scenarios as feature vectors \(\{\mathbf{z}_i\}\). For the current state \(\mathbf{z}_t\in\mathbb{R}^d\), we retrieve the most similar scenario \(\mathbf{z}_i\) via cosine similarity, thus leveraging outcomes of analogous past experiences to guide decision-making:
\begin{equation}
    \text{sim}(\mathbf{z}_t,\mathbf{z}_i) 
= \frac{\mathbf{z}_t \cdot \mathbf{z}_i}{\|\mathbf{z}_t\|\|\mathbf{z}_i\|}
\end{equation}

Building on these real-time and historical insights, the engine constructs a comprehensive prompt \(\mathcal{P}_t\) that integrates TTCP-derived risks, scenario-specific experiences, and traffic knowledge. This prompt includes road geometry, vehicle positions, and conflict zones, along with domain heuristics to provide the LLM with a rich contextual foundation for action proposals. To enhance reliability and minimize hallucinations, we then employ a chain-of-thought approach ~\cite{wei2022chain} in which the model iteratively evaluates collision severity, short- and long-term maneuver consequences, and broader traffic implications. This structured reasoning process reduces logical inconsistencies, resulting in safer and more interpretable autonomous driving policies.

\begin{algorithm}
    \label{LLM Teacher}
    \SetAlgoLined
    \SetKwInOut{Input}{Input}\SetKwInOut{Output}{Output}
    \caption{LLM Teacher Agent} 
    
    \Input{State $s(t)$, Memory $\mathcal{M}$}
    \Output{Action $a_t \in A$}
    
    \For{$t \leftarrow 0$ \KwTo $T_{max}$}{
        
        \For{$i \leftarrow 1$ \KwTo $N_{vehicle}$}{
            \textbf{Identify critical risks:} \\
            \begin{small}
                $\tau_i = \arg\min_{\Delta t \ge 0} \|\mathbf{p}_\text{ego}(t + \Delta t) - \mathbf{p}_i(t + \Delta t)\|$
            \end{small} \\
            $\tau_{\min} \leftarrow \min \{\tau_1,\dots,\tau_N\}$ \\
        }
        
        \textbf{Construct Current State Vector:} \\
        \begin{small}
            $\mathbf{z}_t \leftarrow \bigl(\mathbf{p}_\text{ego}(t), \mathbf{v}_\text{ego}(t), \{\mathbf{p}_i(t), \mathbf{v}_i(t)\}_{i=1}^N, \{\tau_i\}_{i=1}^N\bigr)$
        \end{small} \\
        
        \textbf{Memory retrieval with cosine similarity:} \\
        $\hat{j} \leftarrow \arg\max_{1 \le j \le M} \frac{\mathbf{z}_t \cdot \mathbf{z}_j}{\|\mathbf{z}_t\|\|\mathbf{z}_j\|}$ \\
        Construct prompt $\mathcal{P}_t \leftarrow \text{Retrieval}\Bigl(\mathbf{s}_t, \mathcal{M}\Bigr)$\\

        \textbf{CoT Reasoning:} \\
        $\mathcal{A}_t \leftarrow \{\text{Decoding LLM final decision}\}$ \\    
        
        \textbf{Reflective Evaluator Update:} \\
        Calculate risk: $\Omega: \mathcal{S} \times \mathcal{A} \rightarrow \mathbb{R}^+$ \\
        \If{$\max_{t} \Omega(s_t, a_t) \ge \delta$}{
            \textbf{Further Reflection:} \\
            \begin{small}
                $\bigl\{\Delta{Policy},\; \Delta{Prompt},\; \Delta{Constraint}\bigr\} \;\leftarrow\; f_{LLM}\bigl(\mathcal{Q}_{ref}\bigr)$
            \end{small}
        }
        $\mathcal{M} \leftarrow \text{Updated Memory after Reflection}$  
    }
    \textbf{return} $a_t$, $\mathcal{M}$
\end{algorithm}

\subsubsection{\textbf{Memory Repository}}
The Memory Repository stores and manages all pertinent knowledge required by the LLM Teacher. It operates as a dynamic database \(\mathcal{M}\) that contains the prior scenarios and policies, which includes the historical states\(\{\mathbf{s}_i\}\), actions\(\{\mathbf{a}_i\}\) and consequences\(\{\mathbf{r}_i\}\).

When generating a new prompt \(\mathcal{P}_t\), the Decision Engine queries \(\mathcal{M}\) for relevant context, ensuring the LLM has immediate access to historical examples and domain constraints. By selectively retrieving and embedding these elements, the LLM Teacher can provide more accurate and context-sensitive guidance:
\begin{equation}
\label{mem_ret}
    \mathcal{P}_t \leftarrow \text{Retrieval}\Bigl(\mathbf{s}_t, \mathcal{M}\Bigr)
\end{equation}

Periodic updates to \(\mathcal{M}\) occur based on newly encountered scenarios or reflective feedback from previous driving sessions. This design allows the LLM Teacher to accumulate knowledge over time, enabling improved reasoning and continuous evolution across diverse driving environments.

\subsubsection{\textbf{Reflective Evaluator}}
The Reflective Evaluator systematically reviews driving episodes to improve decision-making by identifying risky events and integrating learned lessons into future policies.

After each driving session, we first collect the experience tuples:
\begin{equation}
    \mathcal{D} = \{(s_t, a_t, s_{t+1}) \mid t = 1,2,\dots,T\}
\end{equation}
where \(s_t\) and \(a_t\) denote the state and action at time \(t\) and \(s_{t+1}\) denotes the subsequent state. To pinpoint high-risk segments, we define a risk function \(\Omega: \mathcal{S} \times \mathcal{A} \rightarrow \mathbb{R}^+\) that quantifies the potential for collisions or other undesirable outcomes.
\begin{equation}
    \Omega(s_t, a_t) \;=\; \max\Bigl(\tfrac{1}{\tau_{TTCP}(s_t, a_t)}, \;\beta\,\mathbb{I}\{infraction\}\Bigr)
\end{equation}
where \(\tau_{TTCP}(s_t, a_t)\) is the TTCP for action \(a_t\) in state \(s_t\), \(\mathbb{I}\{\cdot\}\) is an indicator function for specific infractions, and \(\beta\) is a weighting constant.Any episode with \(\max_{t}\Omega(s_t, a_t)\ge\delta\) is flagged for further reflection.

For the flagged segment \(\{(s_i,a_i)\}_{i=k}^m\), the LLM is prompted to analyze the sequence of risky actions and causes. Through CoT reasoning, it proposes a domin-specific adjustment:
\begin{equation}
\bigl\{\Delta{Policy},\; \Delta{Prompt},\; \Delta{Constraint}\bigr\} 
\;\leftarrow\; f_{LLM}\bigl(\mathcal{Q}_{ref}\bigr)
\end{equation}

These updated constraints and policies are then integrated back into the memory repository \(\mathcal{M}\) and the decision engine's prompt construction logic. By iterating this reflection process, the LLM-Teacher systematically reduces error recurrence and strengthens overall policy robustness.

\subsection{DRL student}
\subsubsection{\textbf{Actor-Critic Algorithm with Policy Constraint}}
An actor-critic framework\cite{grondman2012survey} is adopted, where both the state-value function \(V^\pi\) and the action-value function \(Q^\pi\) are recursively estimated. For a policy \(\pi(\mathbf{a}\mid\mathbf{s})\), the state-value function and the corresponding action-value function at state \(\mathbf{s}_t\) is:
\begin{equation}
V^{\pi}(\mathbf{s}_{t}) 
= \underset{\mathbf{a}_{t}\sim\pi(\cdot \mid \mathbf{s}_{t})}{\mathbb{E}}\Bigl[\,Q^{\pi}\bigl(\mathbf{s}_{t},\mathbf{a}_{t}\bigr)\Bigr]
\end{equation}

\begin{equation}
Q^{\pi}\bigl(\mathbf{s}_{t},\mathbf{a}_{t}\bigr)
=r(\mathbf{s}_{t},\mathbf{a}_{t}) 
+ \gamma \,\mathbb{E}_{\mathbf{s}_{t+1}\sim T(\cdot \mid \mathbf{s}_{t},\mathbf{a}_{t})}\bigl[V^{\pi}(\mathbf{s}_{t+1})\bigr]
\end{equation}

The goal of the algorithm is to determine the optimal policy \(\pi^{*}\) that maximizes \(V^{\pi}(\mathbf{s})\) for all \(\mathbf{s}\in\mathcal{S}\). In our proposed algorithm, we iteratively learn the V function and Q function by minimizing the mean-squared Bellman error (MSBE) and optimize the policy \(\pi\) by maximizing the Q value, where MSBE is defined as:
\begin{small}
    \begin{equation}
\mathcal{L}(\phi_i)
=\mathbb{E}_{\bigl(\mathbf{s}_t,\mathbf{a}_t,r_t,\mathbf{s}_{t+1}\bigr)\sim\mathcal{B}}
\Bigl[
    \bigl(
        Q_{\phi_i}(\mathbf{s}_t,\mathbf{a}_t)
        - \bigl(r_t+\gamma\,V_{g}(\mathbf{s}_{t+1})\bigr)
    \bigr)^2
\Bigr]
\end{equation}
\end{small}
where \(\mathcal{B}\) is the experience replay buffer, and \(V_{g}\) represents a periodically updated target value function. The actor network is optimized by selecting actions \(\mathbf{a}_t\) that maximize the critic’s estimate \(Q_{\phi_i}\bigl(\mathbf{s}_t,\mathbf{a}_t\bigr)\), thereby promoting higher return.

To incorporate demonstration actions from the LLM Teacher’s policy \(\pi^{T}\) into the actor-critic framework and guide the DRL agent’s policy \(\pi_{S}\) during early exploration, we introduce a Kullback–Leibler (KL)\cite{huang2022efficient} divergence constraint. The agent’s learning objective is formulated as a constrained optimization problem:
\begin{align}
&\min_{\pi^{S}}
\mathbb{E}_{\substack{\mathbf{s}_{t}\sim\mathcal{D}\\\tilde{\mathbf{a}}_{t}\sim\pi^{S}(\mathbf{s}_{t})}}
\bigl[
    -Q_{\phi_{i}}\bigl(\mathbf{s}_{t},\tilde{\mathbf{a}}_{t}\bigr)
\bigr]
\nonumber \\
&\quad\text{s.t.}\quad
\hat{D}_{\mathrm{KL}}
\Bigl(\pi^{S}(\mathbf{s}_{t}),\,\pi^{T}(\mathbf{s}_{t})\Bigr) 
\,\leq\, \sigma
\end{align}
where \(\sigma > 0\) is a tolerance that bounds the KL divergence between the agent’s policy \(\pi^{S}(\mathbf{s}_{t})\) and the teacher’s policy \(\pi^{E}(\mathbf{s}_{t})\). During early training, \(\sigma\) is kept small to enforce proximity to the teacher’s demonstrated actions, thereby accelerating convergence. As training proceeds, \(\sigma\) gradually expands, allowing the agent to rely more on its own exploration while still incorporating early guidance. This procedure balances leveraging teacher knowledge for rapid initial learning with the agent’s intrinsic exploration for robust final performance.

\subsubsection{\textbf{Policy Distillation and Fusion}}

\begin{figure}[htp]
    \centering
    \includegraphics[width=0.9\linewidth]{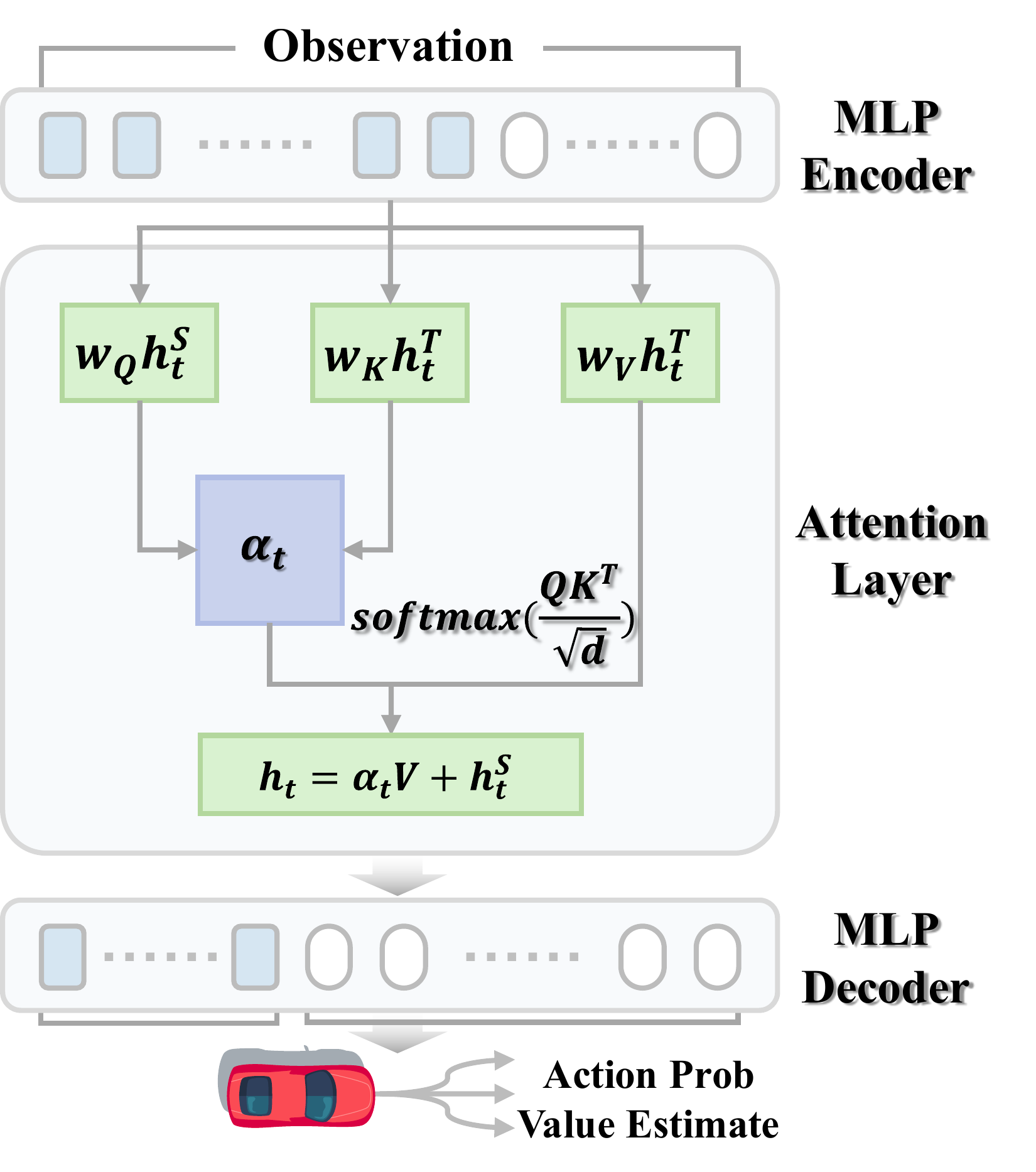}
    \caption{Proposed policy network with self-attention layer. The network integrates self-attention to estimate action probabilities and value functions from the teacher’s strategy, enabling strategy distillation and a balance between teacher guidance and self-exploration.}
    \label{fig:transformer}
\end{figure}

Although the Teacher Agent offers high-level guidance, it does not directly provide action probabilities or value estimates. To bridge this gap, we embed a Transformer-based self-attention mechanism shown in Fig.~\ref{fig:transformer} within the Student’s policy network. This component approximates the Teacher’s implicit policy and fuses it with the Student’s learned strategy in a flexible, data-driven manner.

Let \(\mathbf{s}_t \in \mathcal{S}\) be the state at time \(t\). We introduce two embeddings:
\begin{equation}
    \mathbf{h}_t^S = f_S(\mathbf{s}_t), 
    \quad
    \mathbf{h}_t^T = f_T(\mathbf{s}_t)
\end{equation}
where \(f_S\) and \(f_T\) are neural encoders for the Student and the Teacher, respectively. The vector \(\mathbf{h}_t^T\) is learned to approximate the implicit Teacher policy, \(\hat{\pi}^{T}\), and its corresponding action-value function, \(\hat{Q}^{T}\). For each action \(\mathbf{a}\in\mathcal{A}\):
\begin{align}
\hat{\pi}^T(\mathbf{a}\mid \mathbf{s}_t) 
&= softmax\bigl(\mathcal{W}_{p}\,\mathbf{h}_t^T + \mathbf{b}_{p}\bigr)\\
\hat{Q}^T(\mathbf{s}_t,\mathbf{a}) 
&= \mathcal{W}_{q}\,\mathbf{h}_t^T + \mathbf{b}_{q}
\end{align}
where \(\{\mathcal{W}_p, \mathcal{W}_q, \mathbf{b}_p, \mathbf{b}_q\}\) are learnable parameters.

To integrate these dual embeddings, a self-attention mechanism is employed:
\begin{equation}
    Q = \mathcal{W}_Q\,\mathbf{h}_t^S, 
    \quad
    K = \mathcal{W}_K\,\mathbf{h}_t^T,
    \quad
    V = \mathcal{W}_V\,\mathbf{h}_t^T,
\end{equation}

where \(\mathcal{W}_Q, \mathcal{W}_K, \mathcal{W}_V\) are learnable projections. The self-attention coefficient \(\alpha_t\) can be described as:
\begin{equation}
    \alpha_t 
    = softmax \Bigl(
    \frac{QK^\top}{\sqrt{d}}
    \Bigr)
\end{equation}
with \(d\) denoting the dimensionality of \(K\). The fused representation \(\mathbf{h}_t\) is:
\begin{equation}
\mathbf{h}_t 
= \alpha_t \,V + \mathbf{h}_t^S 
= \alpha_t \,\mathcal{W}_V\,\mathbf{h}_t^T + \mathbf{h}_t^S
\end{equation}

In multi-head settings, the process is replicated across several attention heads, and the outputs are concatenated.

The Student uses the fused embedding \(\mathbf{h}_t\) to produce its final policy \(\tilde{\pi}\) and action-value estimate \(\tilde{Q}\):
\begin{align}
\tilde{\pi}(\mathbf{a}\mid \mathbf{s}_t) 
&= softmax\bigl(\mathbf{W}_{\pi}\,\mathbf{h}_t + \mathbf{b}_{\pi}\bigr)\\
\tilde{Q}(\mathbf{s}_t,\mathbf{a})
&= \mathbf{W}_{Q}\,\mathbf{h}_t + \mathbf{b}_{Q}
\end{align}

The self-attention parameters and teacher embeddings are optimized jointly. If demonstration data \(\{(s,\mathbf{a}_T)\}\) is available, an auxiliary distillation loss enforces consistency with the Teacher’s decisions:
\begin{equation}
\mathcal{L}_{\mathrm{distill}} 
= -\,\mathbb{E}_{(s,\mathbf{a}_T)\in \mathcal{D}_T}
\bigl[\log\,\hat{\pi}^T(\mathbf{a}_T\mid s)\bigr]
\end{equation}

This term encourages \(\mathbf{h}_t^T\) to distill excellent policies from the demonstrated behavior of the LLM Teacher, while the Student continues to learn its own policy through exploration and reward feedback.

\section{Simulation and Performance Evaluation}
\subsection{Driving scenarios}
\begin{figure}[htp]
    \centering
    \includegraphics[width=0.9\linewidth]{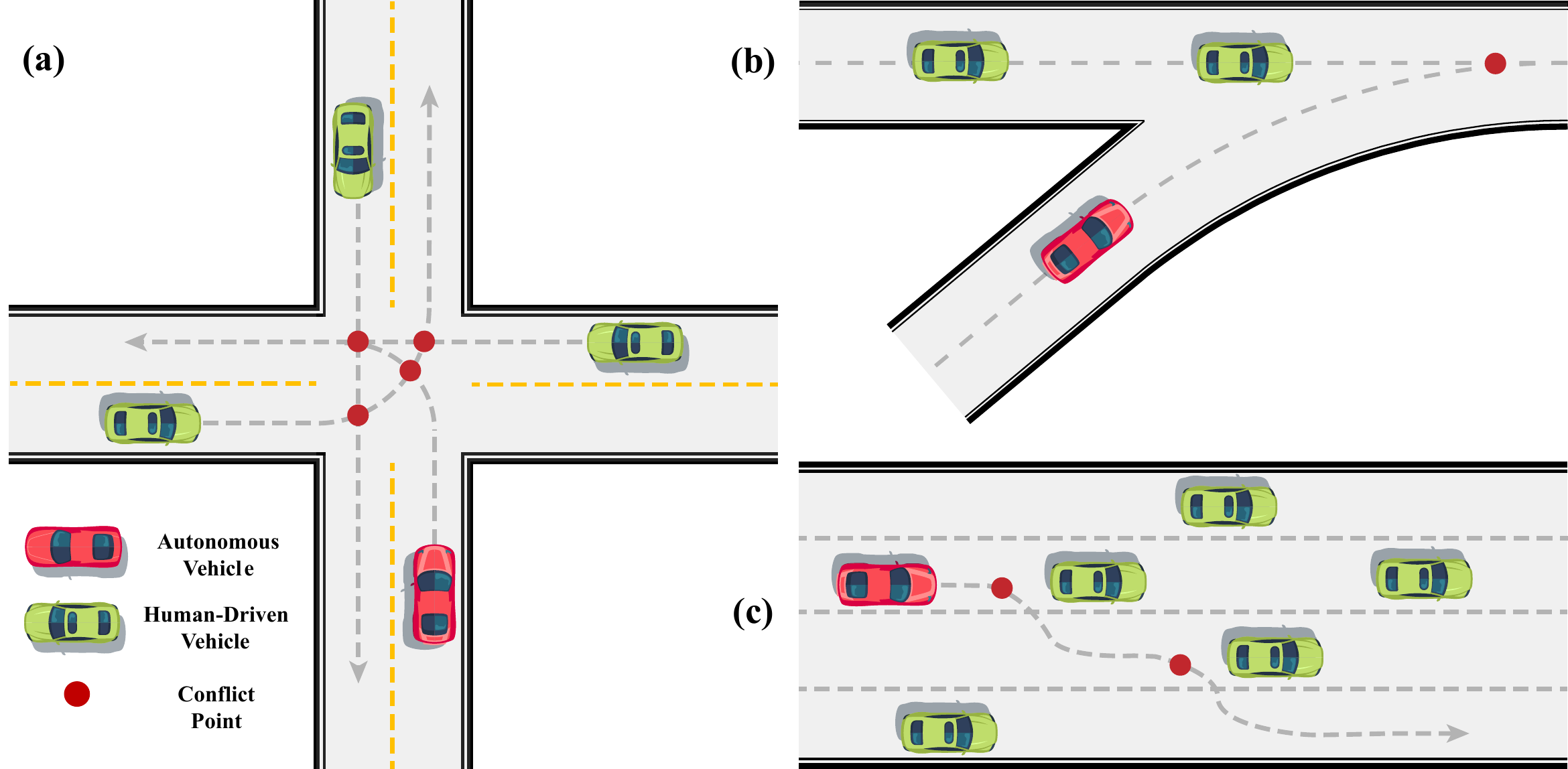}
    \caption{The designed gradient verification scenario for simulation: (a) Unsignalized Intersection; (b) High-Speed Ramp Merging; (c) Four-Lane Adaptive Cruise.}
    \label{fig:experimental_sce}
\end{figure}

We evaluate the comprehensive performance of our autonomous driving model using a gradient verification scenario constructed with Highway-Env\cite{highway-env} and OpenAI Gym. To capture a broad spectrum of driving complexities, we design three heterogeneous task systems with progressively decreasing difficulty, as illustrated in Fig.~\ref{fig:experimental_sce}:  
\begin{enumerate}
    \item Unsignalized intersection (Fig.~\ref{fig:experimental_sce}(a)): The agent must execute an unprotected left turn at an unsignalized intersection, requiring conflict resolution and time-slot preemption to navigate crossing traffic safely.  
    \item High-Speed Ramp Merging (Fig.~\ref{fig:experimental_sce}(b)): The agent operates on an acceleration lane, performing speed matching and gap selection to merge seamlessly into highway traffic at elevated velocities.  
    \item Four-Lane Adaptive Cruise (Fig.~\ref{fig:experimental_sce}(c)): The agent focuses on fine-grained control of inter-vehicle distances and speeds across four lanes, highlighting precision in longitudinal control and continuous lane tracking.
\end{enumerate}

By varying parameters, we simulate \emph{conservative}, \emph{standard}, and \emph{aggressive} driver profiles, each featuring different desired speeds, accelerations, and tolerances for spacing. Vehicle speeds follow a normal distribution centered within a reasonable range, and we introduce a 15\% abnormal speed disturbance to emulate real-world deviations.

\subsection{Implementation Details}
Both our model and the baseline methods utilize a policy network composed of a multilayer perceptron (MLP) with two hidden layers of size \(128 \times 128\). We employ two self-attention heads, each also of dimension 128, to fuse the Student and Teacher representations.The clip range is dynamically adjusted using a linear schedule, starting with an initial value and decreasing according to the remaining training progress. Each model is trained for at least \(10^5\) time steps, with an evaluation performed every 500 time steps. We use \textit{GPT-4o-mini}\cite{hurst2024gpt} as our LLM backbone, which shows reliable logical reasoning and real-time decision-making in driving tasks; it serves as the Teacher Agent for only the first 10\% of training steps, after which constraints are gradually relaxed to encourage independent exploration. The specific parameter settings are shown in Table \ref{parameter}. All experiments run on a computing platform equipped with Intel(R) Core(TM) i7-14700K CPU, an NVIDIA GeForce RTX 4080 SUPER GPU and 32 GB of RAM.

\begin{table}[ht]
\centering
\caption{Hyperparameters used in the experiment}
\label{parameter}
\begin{tabular}{ccc}
\toprule
\textbf{Symbol} & \textbf{Meaning} & \textbf{Value} \\
\midrule
$\alpha$ & Learning rate & $5\times10^{-4}$ \\
$\mathcal{N}_{train}$ & Minimum total training steps & $1\times10^{5}$ \\
$\gamma$ & Discount factor & 0.99 \\
$\epsilon$ & Initial value of clip range & 0.2 \\
$\mathcal{B}$ & Training batch size & 128 \\
$\mathcal{N}_{B}$ & Rollout buffer size & 1600 \\
$\|\mathcal{M}\|$ & Capacity of Memory Repository & 20 \\
$\mathcal{N}_{shot}$ & Number of examples for few-shot learning & 3 \\
\bottomrule
\end{tabular}
\end{table}

\subsection{Performance Evaluation}
\subsubsection{Comparison with Baseline Methods}

\begin{figure*}
    \centering
    \includegraphics[width=\linewidth]{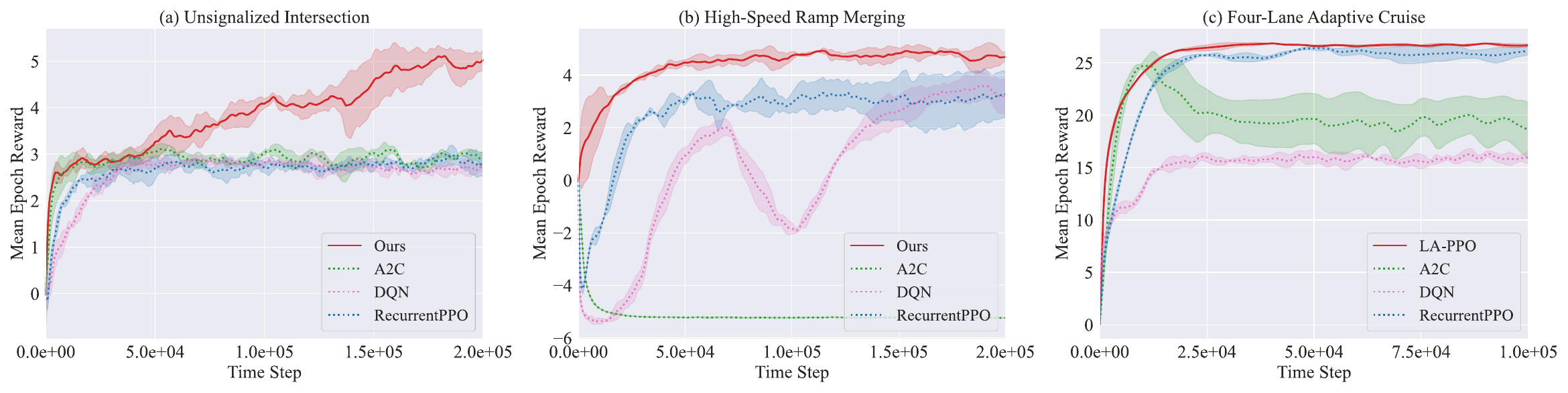}
    \caption{Comparison of the performance of this model with traditional DRL training results.}
    \label{fig:baselines}
\end{figure*}

\begin{figure*}
    \centering
    \includegraphics[width=\linewidth]{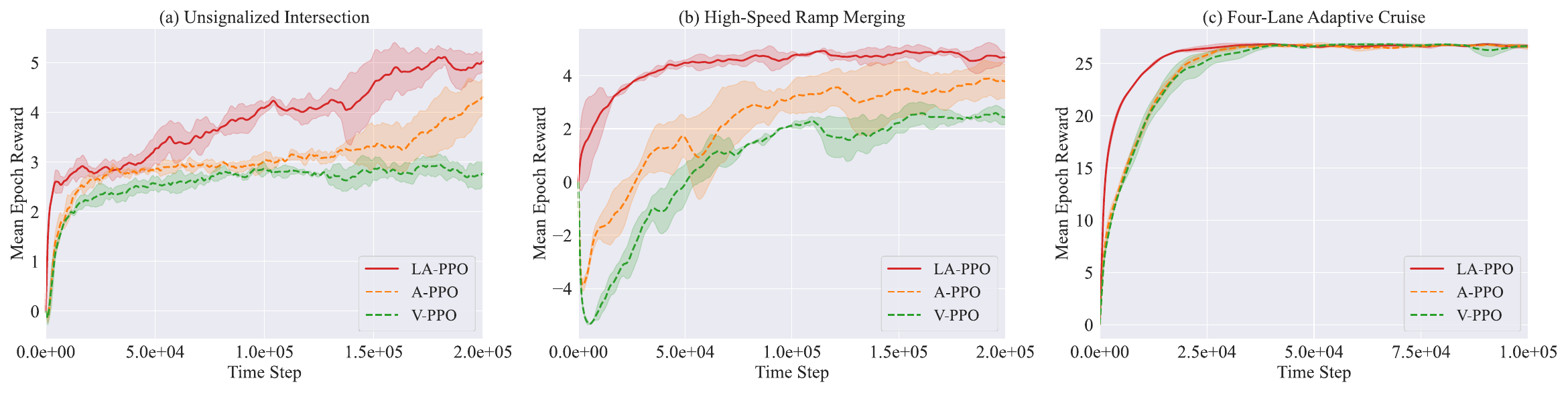}
    \caption{Comparison of performance results during the ablation experiment training process.}
    \label{fig:ablation}
\end{figure*}


To assess the effectiveness of our approach, we benchmark four algorithms: a value-based method (DQN\cite{mnih2013playing}), a policy-gradient method (A2C\cite{mnih2016asynchronous}), a sequence-memory-based method (RecurrentPPO\cite{pleines2022generalization}), and the current LLM-based state-of-the-art (Dilu\cite{wen2023dilu}). We record the average return during training in Fig.~\ref{fig:baselines}, where the solid lines denote mean performance and the shaded regions indicate 95\% confidence intervals.

In unsignalized intersection shown in Fig.~\ref{fig:baselines}(a), Our model rapidly improves during the initial training phase and converges to the highest final return. A2C exhibits significant fluctuations, implying instability near convergence. While other baselines eventually stabilize, their end-stage returns remain notably below ours, highlighting a substantial performance gap.
In high-speed ramp merging shown in Fig.~\ref{fig:baselines}(b), our method achieves high returns early on, stabilizing around \(5\times 10^{4}\) steps and consistently maintaining near-optimal performance thereafter. In contrast, DQN starts with negative returns and steadily climbs to a suboptimal plateau. A2C fares the worst, likely owing to its sensitivity in time-critical merging tasks. Although RecurrentPPO converges more promptly than A2C, its ultimate reward remains below our model’s, underscoring the challenges of handling highly dynamic traffic with simple recurrent mechanisms.
All methods experience rapid early gains, yet differ significantly in final returns and stability in four-lane adaptive cruise shown in Fig.~\ref{fig:baselines}(c). Our model maintains a leading position throughout and converges to a near-maximal reward. RecurrentPPO is intermittently competitive but prone to fluctuations. DQN and A2C both show moderate terminal performance, with A2C stabilizing late but still achieving a lower reward ceiling.

What's more, Table~\ref{tab:results} provides a numerical summary of success rate, evaluation return, average speed, \(\Delta\)TTCP, and decision-making time for each approach. In the unsignalized intersection scenario, our method attains the highest success rate (88\%) while balancing speed and safety margins. For high-speed ramp merging, it achieves 91\% success and outperforms the baselines in average return. Notably, A2C, despite having the highest speed, completely fails (0\% success), demonstrating that overly aggressive driving sacrifices safety and thus overall performance. In four-lane adaptive cruise, our method reaches a perfect success rate (100\%) alongside near-optimal speed and return. Although Dilu shows better speed and safety margins than traditional DRL methods, its extended reasoning time limits online deployment. Overall, our approach surpasses both conventional DRL algorithms and the LLM-based Dilu, underlining its effectiveness and robustness across diverse scenarios.

\subsubsection{Ablation Study}

To evaluate the impact of each component, we conduct an ablation study comparing \emph{Vanilla PPO} (V-PPO\cite{schulman2017proximal}), \emph{Attention-based PPO} (A-PPO), and our \emph{LLM-Guided Attention PPO} (LA-PPO). As shown in Fig.~\ref{fig:ablation}, LA-PPO demonstrates faster convergence and higher final rewards relative to V-PPO and A-PPO, indicating superior stability and robustness. In the more demanding scenarios,  such as unsignalized intersection and high-speed ramp merging, LA-PPO quickly attains higher returns and preserves its advantage throughout training. Although all methods converge to similar rewards in the simpler four-lane adaptive cruise task, LA-PPO still displays a slight edge in convergence rate and peak performance. These observations confirm that leveraging LLM guidance in conjunction with an attention mechanism yields more effective teacher knowledge transfer and better-directed policy learning.

\subsubsection{Teacher-Student Comparison}
\begin{figure}[htp]
    \centering
    \includegraphics[width=\linewidth]{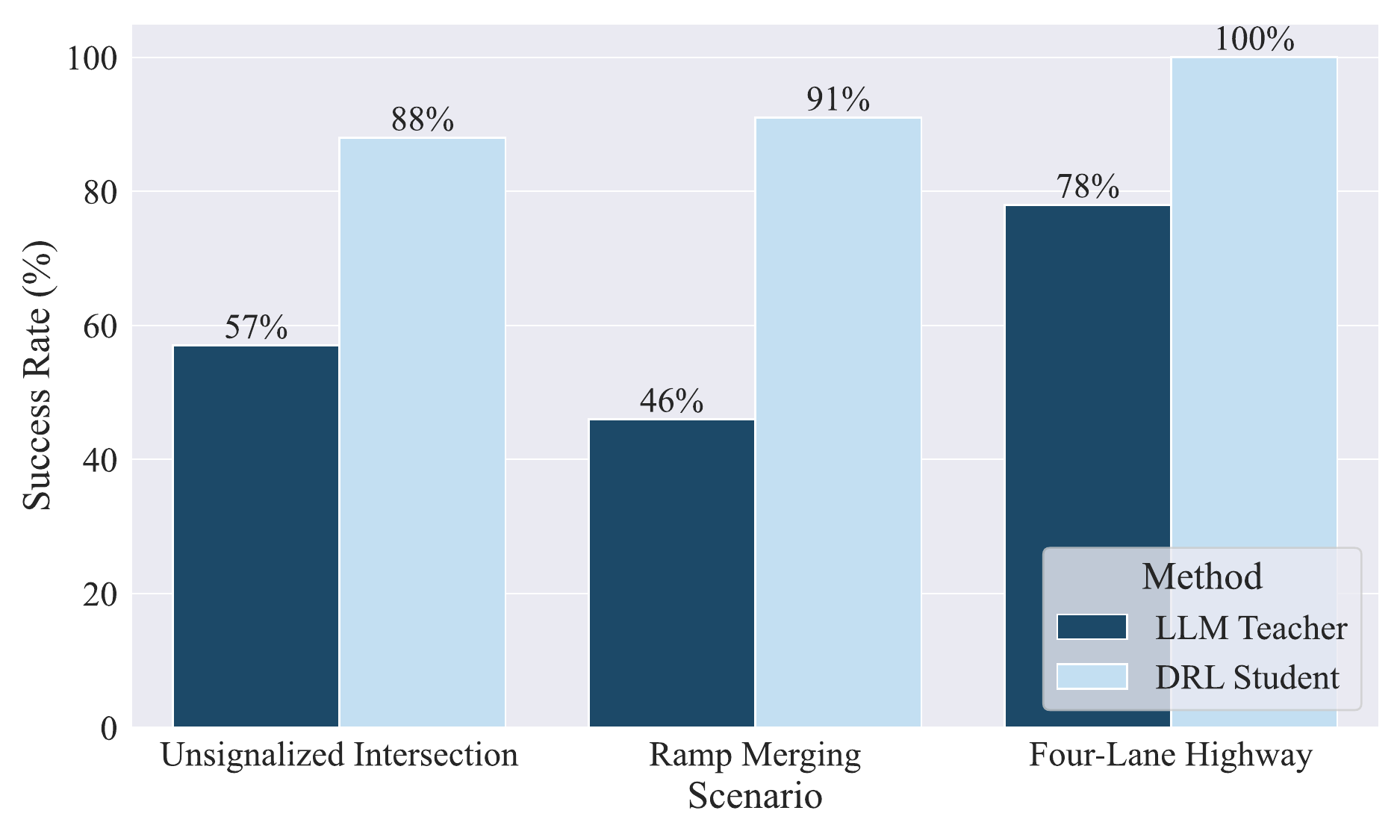}
    \caption{Comparison of testing success rate results between the teacher agent and the student agent.}
    \label{fig:success_rate_comparison}
\end{figure}

As shown in Fig.~\ref{fig:success_rate_comparison}, we further examine success rates for the LLM Teacher and the DRL Student in each scenario. The Student outperforms the Teacher across all tasks, illustrating that while LLM guidance aids rapid early-stage learning, continual environment interaction empowers the Student to refine and ultimately surpass the Teacher’s performance. This outcome highlights the strengths of a teacher-student paradigm in autonomous driving policy learning.

\subsection{Case Analysis}

\begin{table*}
    \centering
    \caption{Comparison of security, efficiency, and real-time test results of different methods in multiple scenarios.}
    \label{tab:results}
    \begin{tabular}{llc ccc c}
        \toprule
        Sce & Model & Success Rate (\%) & Eval Reward & Avg. Speed (m/s) & $\Delta$TTCP (s) & Consumption Time (s) \\
        \midrule
                     & DQN & 58 & 3.22 & 8.77 & 5.22 & 0.002 \\
                     & A2C & 54 & 2.88 & 9.02 & 5.05 & 0.003 \\
        Intersection & RecurrentPPO & 74 & 0.86 & 5.34 & 4.98 & 0.003 \\
                     & Dilu & 57 & 6.59 & 7.32 & 1.53 & 3.906 \\
                     & Ours & \textbf{88} & \textbf{5.68} & 7.36 & \textbf{4.92} & \textbf{0.004} \\
        \midrule
               & DQN & 83 & 2.81 & 13.01 & 1.21 & 0.002 \\
               & A2C & 0 & -5.20 & 28.28 & 0.37 & 0.002 \\
        Merge  & RecurrentPPO & 30 & 1.83 & 22.05 & 0.28 & 0.003 \\
               & Dilu & 46 & 3.21 & 14.92 & 0.72 & 6.166 \\
               & Ours & \textbf{91} & \textbf{5.60} & 16.50 & \textbf{1.31} & \textbf{0.004} \\
        \midrule
                 & DQN & 71 & 22.02 & 21.96 & 1.96 & 0.002 \\
                 & A2C & 50 & 21.95 & 25.00 & 2.18 & 0.003 \\
        Highway  & RecurrentPPO & 88 & 25.55 & 20.30 & 2.18 & 0.003 \\
                 & Dilu & 78 & 29.53 & 23.39 & 2.21 & 6.131 \\
                 & Ours & \textbf{100} & \textbf{27.17} & 23.53 & \textbf{2.13} & \textbf{0.003} \\
        \bottomrule
    \end{tabular}
\end{table*}

\begin{figure*}
    \centering
    \includegraphics[width=\linewidth]{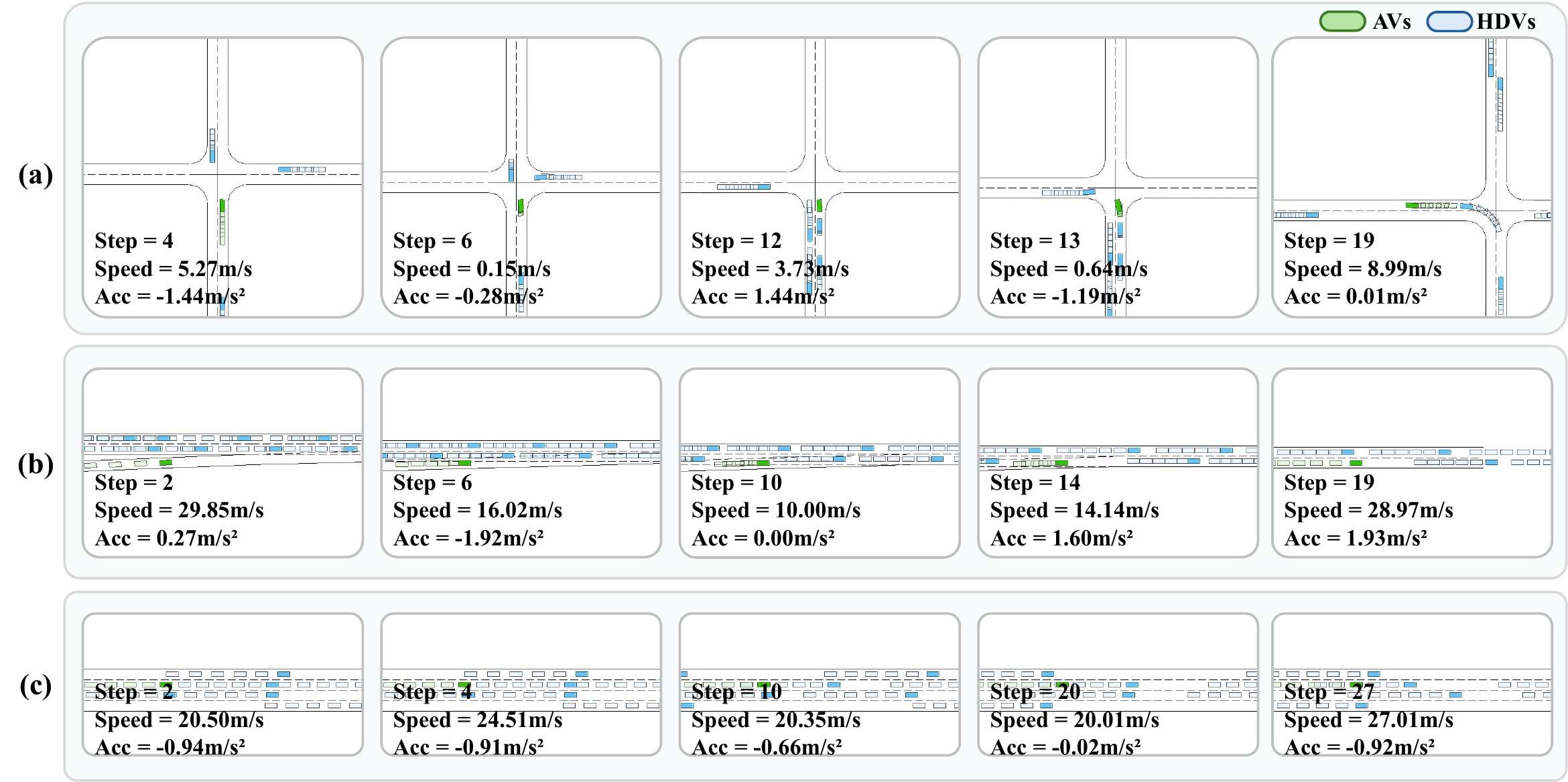}
    \caption{Test case performance results of TeLL-Drive in three scenarios (a) Unsignalized Intersection, (b) High-Speed Ramp Merging, (c) Four-Lane Adaptive Cruise, where green represents AVs guided by TeLL-Drive and blue represents HDVs. } 
    \label{fig:case-analysis}
\end{figure*}

To further explore the decision-making process and behavior characteristics of the proposed model in actual scenarios, we selected three representative cases, as shown in Fig.~\ref{fig:case-analysis}.

In the unsignalized intersection shown in Fig.~\ref{fig:case-analysis}(a), after the agent approaches the stop line of the intersection in step 4, it actively slows down to give way to other vehicles that arrive at the intersection first until step 6; when trying to accelerate again in step 12, it promptly observes that the vehicle in the adjacent lane is about to pass, quickly judges and slows down again, and accelerates to leave after it passes. This behavior fully reflects the model's understanding and compliance with traffic rules and social interactions, and can achieve safe and reasonable interactions with other traffic entities.

In the ramp merging scenario illustrated as Fig.~\ref{fig:case-analysis}(b), the model first accelerates quickly to complete the merging action; in step 6, it actively slows down to maintain a safe distance from the vehicle in front, and accelerates again after confirming that there is enough safe distance between it and the vehicle in front in step 14, and finally merges smoothly into the main road. This process shows that the agent has precise control over the acceleration and deceleration decisions in high-speed scenarios and a keen perception of risk factors.

In the example of four-lane adaptive cruise control shown in Fig.~\ref{fig:case-analysis}(c), the agent can continuously monitor and maintain a safe distance from the vehicle in front in dense traffic conditions or even in the presence of traffic disturbances, and adjust the speed in a timely manner to avoid rear-end collisions or excessive deceleration. This case shows that the model has good stability and active safety in long-term cruise tasks.

From the above cases, it can be seen that our model can demonstrate good interaction capabilities and strategic decision-making levels in a variety of complex driving scenarios, which further supports the conclusions of the aforementioned quantitative experiments.


\section{Vehicle-in-Loop Experiment}
\subsection{Virtual-Real Integration Experimental Platform}

To further assess the robustness and real-time performance of TeLL-Drive, we conduct a vehicle-in-loop experiment that combines virtual and real-world testing. A fusion platform is developed to integrate virtual traffic simulations with real vehicle hardware, allowing for the evaluation of the intelligent driving function in dynamic and complex traffic environments. This experimental setup enables the testing of autonomous driving decision-making under various conditions, including scenarios with potential safety hazards, both in virtual and real-world settings.

\begin{figure}
    \centering
    \includegraphics[width=\linewidth]{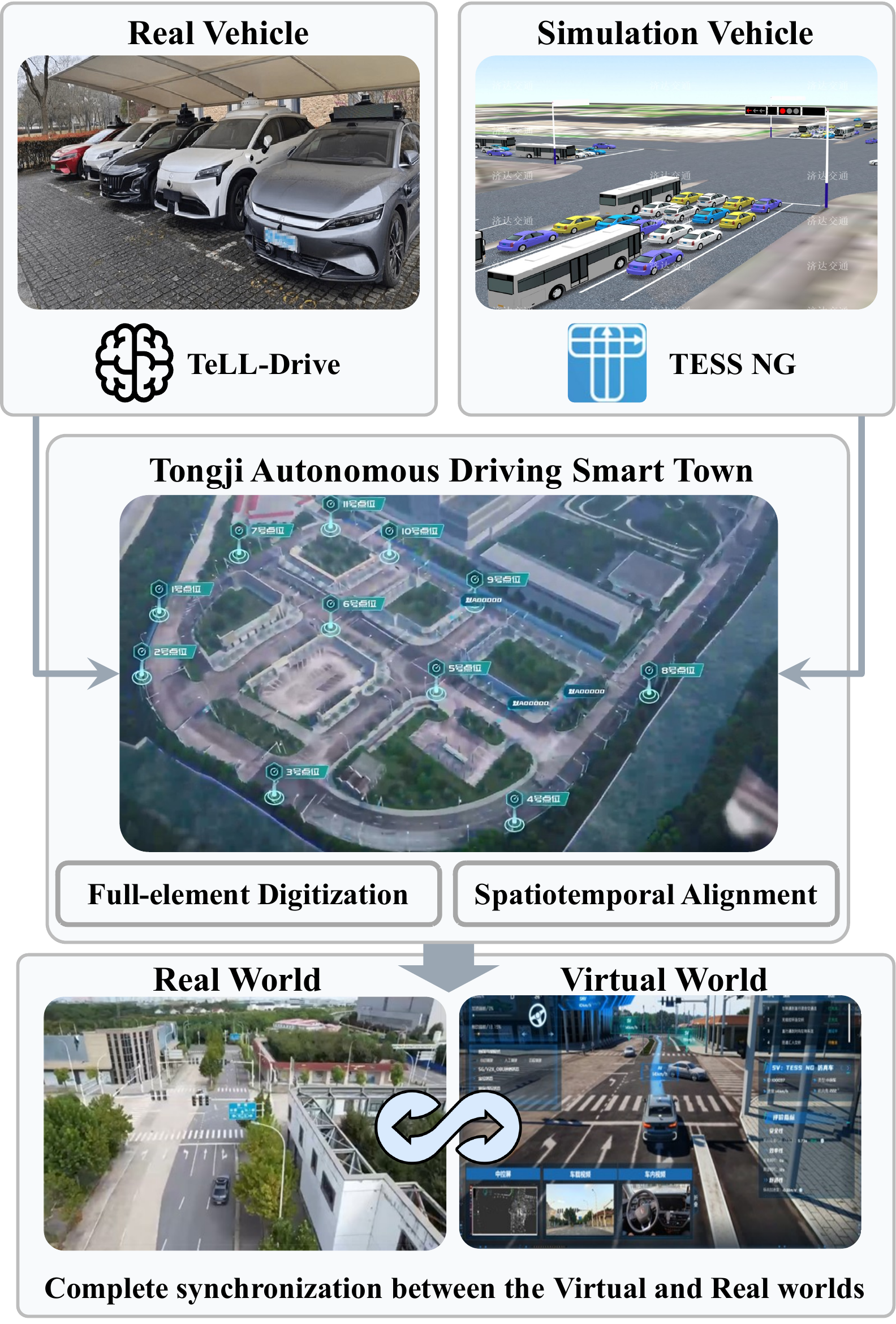}
    \caption{The virtual-reality fusion experimental platform built based on Tongji Smart Town, which achieves complete synchronization between the virtual and real worlds. The autonomous driving vehicle uses the TeLL-Dirve decision algorithm, and the virtual vehicle uses TESSNG simulation operation.}
    \label{fig:platform}
\end{figure}

The virtual-real fusion platform consists of two main components: the AV hardware and traffic flow simulation software. As shown in Fig.~\ref{fig:platform}, the traffic flow simulation software generates a virtual traffic environment, providing background traffic data that interacts with the real-world data captured by the AV's sensors. These sensors collect real-time environmental information, which is then fused with the simulated traffic data through a data fusion process. This combined perception is transmitted to the planning control unit, which uses it to generate the vehicle’s motion trajectory. The resulting vehicle trajectory is then fed back into the simulation software, allowing for interaction between the AV and the virtual traffic flow.

In this experiment, the intelligent agent trained under the TeLL-Drive framework serves as the decision-making algorithm for the autonomous vehicle. The simulation vehicle operates using TESSNG\cite{kang2020modeling}, a high-level microscopic simulation software, which enables detailed modeling of vehicle dynamics in complex traffic scenarios. The experiment was conducted at Tongji University’s Autonomous Driving Smart Town, with the unprotected left turn at a complex intersection chosen as the test scenario. As the scenario with the lowest success rate in the simulation experiment, this scenario has inherent safety risks and requires precise decision-making in a dynamic environment. The integration of high-precision maps, precise timing positioning and full-element digitization enable complete synchronization between the real-world and virtual-world, ensuring that both environments are in sync during testing.
This vehicle-in-loop setup provides a comprehensive platform for evaluating the performance of TeLL-Drive in real-time, dynamic driving scenarios. 

\subsection{Case Study Analysis}
\begin{figure*}
    \centering
    \includegraphics[width=\linewidth]{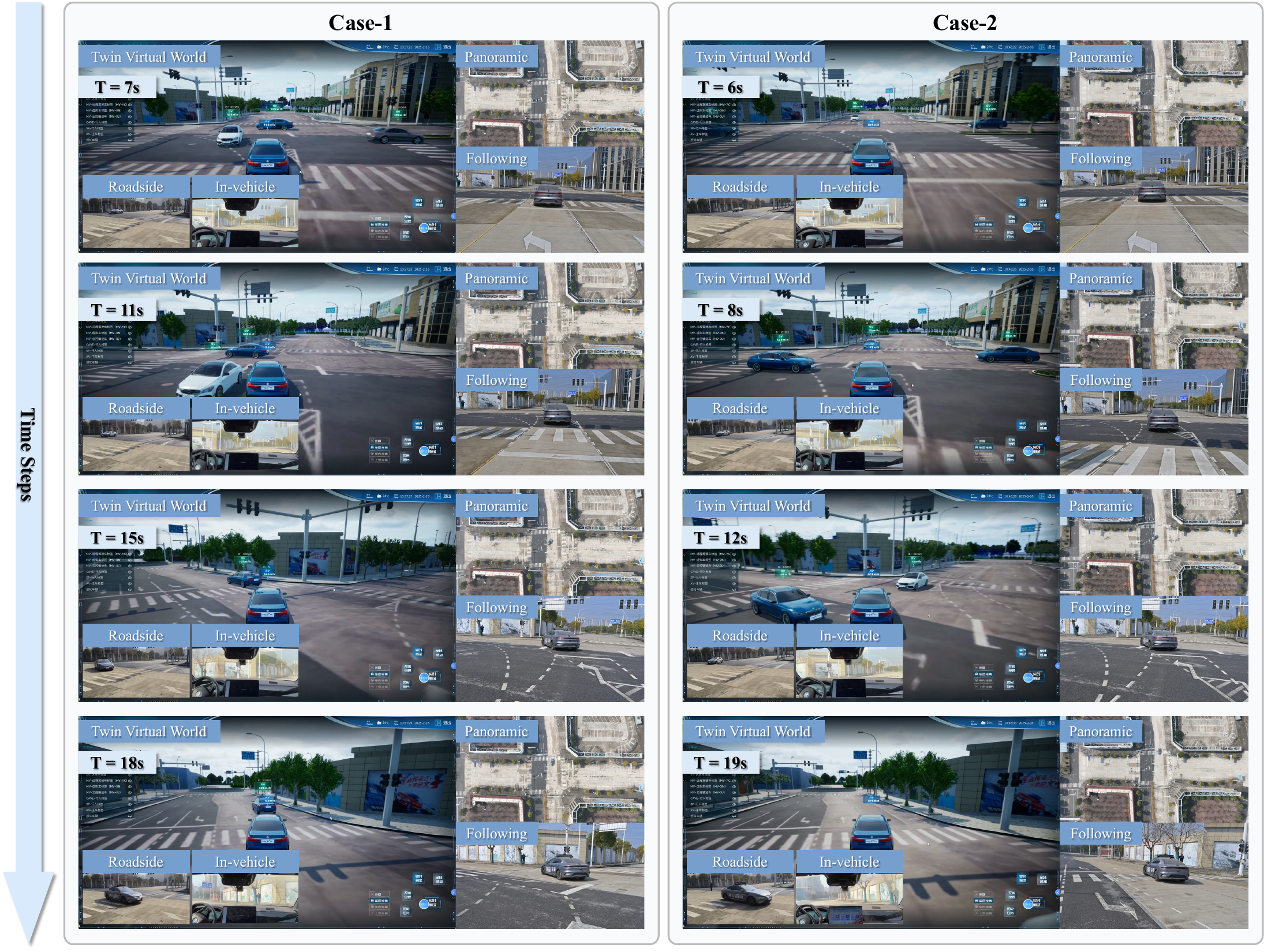}
    \caption{A real vehicle-in-loop experiment based on the virtual-reality fusion experimental platform. The vehicle equipped with TeLL-Drive choose to yield at the intersection in case 1 and has priority in case 2. In both scenarios, vehicles have complex social interactions.}
    \label{fig:real_veh_experiment}
\end{figure*}

We conducted real-vehicle experiments on the virtual-real fusion platform to evaluate the performance of the TeLL-Drive framework. Two representative cases are shown in Fig.~\ref{fig:real_veh_experiment}, each captured from various perspectives, including the virtual twin platform perspective, the drone bird's-eye view, the car-following view, the roadside view, and the in-car perspective. These multiple angles allow for a comprehensive analysis of the algorithm’s performance across different scenarios. The specific experimental video can be accessed on our website\footnote{\href{https://perfectxu88.github.io/TeLL-Drive.github.io/}{Vehicle-in-Loop Experimental Validation Video Weblink}}.

In Case 1, the autonomous vehicle equipped with TeLL-Drive begins from a standstill and accelerates toward the intersection. As it approaches the stop line, the vehicle slows down to create sufficient observation and decision space, enhancing its ability to assess the surrounding traffic. By the 7th second, the vehicle encounters an oncoming vehicle. Upon assessing the situation, the vehicle decides to slow further at the 12th second to yield and avoid a collision. After the oncoming vehicle passes, the autonomous vehicle resumes acceleration and approaches the exit road of the intersection by the 15th second. To maintain a safe distance from the vehicle in front, the system performs adaptive acceleration and deceleration, ensuring both safety and traffic efficiency. In this case, the autonomous vehicle is the last to leave the intersection, but the maneuver was executed safely and efficiently.

In Case 2, the autonomous vehicle follows similar actions up to the point before entering the intersection. However, at the 6th second, the vehicle observes fewer vehicles in the intersection and determines it can pass first, so it accelerates. By the 8th second, a vehicle on the left side approaches, prompting an interaction. After a brief period of strong interaction between the two vehicles, the simulation vehicle (SV) decides to slow down and stop, while our autonomous vehicle continues to pass first, successfully navigating the intersection.

These two cases demonstrate the robustness and reliability of the TeLL-Drive framework when deployed on real vehicles. The system effectively adapts to dynamic traffic scenarios, ensuring safe and efficient decision-making in complex environments. The ability of TeLL-Drive to handle both cooperative and conflict-driven interactions, while maintaining safety and traffic flow, underscores its potential for real-world autonomous driving applications.

\section{Conclusion}
Our proposed TeLL-Drive framework integrates teacher-guided learning with attention-based policy optimization, enabling efficient knowledge transfer and robust decision-making. Experimental results demonstrate that TeLL-Drive outperforms conventional DRL methods and existing LLM-based approaches across multiple metrics, including success rate, average return, and real-time feasibility. Additionally, ablation studies highlight the significance of each model component, particularly the synergy between attention mechanisms and LLM teacher guidance. Finally, vehicle-in-the-loop experiments verify the robustness effectiveness of the model when deployed in practice. These findings confirm that our approach not only accelerates policy convergence but also enhances safety and adaptability across diverse traffic conditions.
In the future, we will explore the application of the TeLL-Drive framework to more dynamic, multi-agent environments and verify its scalability and real-time adaptability through real-vehicle experiments in open road scenarios.

\ifCLASSOPTIONcaptionsoff
  \newpage
\fi

\bibliographystyle{unsrt}
\bibliography{related}

\end{document}